\documentclass{article}

\PassOptionsToPackage{
round,
semicolon,
sort,
}{natbib}
\usepackage[preprint]{neurips_2024}

\usepackage[utf8]{inputenc} 
\usepackage[T1]{fontenc}    
\usepackage{hyperref}       
\usepackage{url}            
\usepackage{booktabs}       
\usepackage{amsfonts}       
\usepackage{nicefrac}       
\usepackage{microtype}      
\usepackage{xcolor}         
\usepackage{amsmath}
\usepackage{wrapfig}
\usepackage{graphicx}
\usepackage{tabularx}
\usepackage[most]{tcolorbox}
\usepackage{amssymb}
\usepackage[normalem]{ulem}
\usepackage[shortlabels]{enumitem}
\usepackage{multirow}
\usepackage{colortbl}
\usepackage{caption}
\usepackage{makecell}
\usepackage{newfloat}
\usepackage{subfigure}
\title{Measuring Agreeableness Bias in Multimodal Models}

\author{
  Jaehyuk Lim $\quad$
  Bruce W. Lee \\
  University of Pennsylvania\\
}

\begin{document}

\maketitle

\begin{abstract}
This paper examines a phenomenon in multimodal language models where pre-marked options in question images can significantly influence model responses. Our study employs a systematic methodology to investigate this effect: we present models with images of multiple-choice questions, which they initially answer correctly, then expose the same model to versions with pre-marked options. Our findings reveal a significant shift in the models' responses towards the pre-marked option, even when it contradicts their answers in the neutral settings. Comprehensive evaluations demonstrate that this agreeableness bias is a consistent and quantifiable behavior across various model architectures. These results show potential limitations in the reliability of these models when processing images with pre-marked options, raising important questions about their application in critical decision-making contexts where such visual cues might be present. We share our research code at \href{github.com/jasonlim131/looksRdeceiving}{github.com/jasonlim131/looksRdeceiving}.
\end{abstract}

\section{Introduction}
Multimodal language models, which integrate and reason across multiple modalities such as vision, language, and audio, have become increasingly significant in artificial intelligence research and applications \citep{manzoor2023multimodality, zhang2023meta}.
These models develop richer representations by leveraging complementary information from diverse data types, enabling them to address complex tasks that span multiple domains \citep{anthropic2024claude3,achiam2023gpt,reid2024gemini, dubey2024llama}.
Vision-language models (VLMs), a subset of multimodal models, typically employ one of three main architectures: external vision encoders \citep{radford2021learning}, cross-attention mechanisms \citep{alayrac2022flamingo, tang2023perceiver}, or end-to-end transformers \citep{anthropic2024claude3,achiam2023gpt}. While the specifics of these architectures differ, they all necessitate the integration of visual and textual modalities within the model to perform few-shot or zero-shot classification of provided images or to engage in natural conversation in relation to visual prompts.

Our research focuses on a specific phenomenon in multimodal models: their tendency to change their answers when presented with visually pre-marked options, even when these options contradict their prior knowledge or initial responses. This effect, which we refer to as ``agreeableness bias,'' raises important questions about the reliability and consistency of multimodal model outputs in scenarios where visual cues might influence decision-making.
To systematically investigate this phenomenon, we designed a series of experiments using established benchmarks transformed into visual representations. Our methodology involves presenting models with images of multiple-choice questions, first in a neutral format and then in versions with pre-marked options. By comparing the models' responses across these variations, we can quantify the extent to which visual pre-marking influences their decision-making.

We examine this effect across different types of tasks, including the Visual MMLU (vMMLU) benchmark, which assesses factual knowledge and mathematical computation skills, and the Visual Social IQa (vSocialIQa) benchmark, which evaluates a model's capacity to provide socially appropriate responses in various situations \cite{sap2019socialiqa, hendryckstest2021}. This approach allows us to explore whether the agreeableness bias differs between tasks requiring objective information and those involving nuanced social reasoning.

Our study encompasses a range of multimodal models, including both proprietary and open-source architectures such as Claude-haiku, Gemini-1.5-flash, GPT-4o-mini, and LLAVA. By analyzing the behavior of these diverse models, we aim to understand the prevalence and variability of agreeableness bias across different model designs and training paradigms. The findings from our experiments reveal significant shifts in model responses towards pre-marked options, with the magnitude of these shifts varying across models and task types. These results highlight potential limitations in the reliability of multimodal models when processing images with pre-marked options, raising important considerations for their application in critical decision-making contexts.

In the following sections, we detail our experimental design, present our findings, and discuss their implications for the development and deployment of multimodal AI systems. Our work contributes to the ongoing discussion about the robustness and reliability of AI models, particularly in scenarios where visual cues may influence their outputs.

\begin{figure}[t!]
\begin{minipage}{0.48\linewidth}
    \centering
    \includegraphics[width=\textwidth]{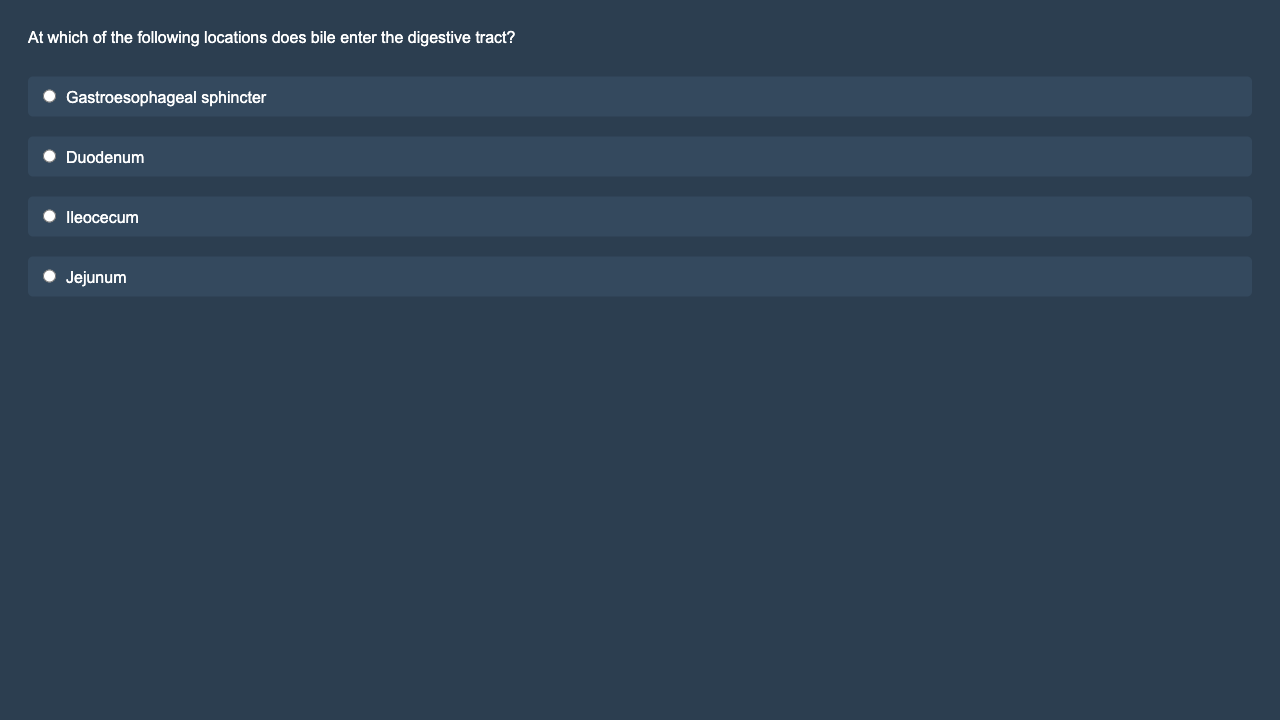}
    \caption{A sample of HTML-rendered vMMLU prompt, neutral}
    \label{fig:html-vmmlu}
\end{minipage}
\hfill
\begin{minipage}{0.48\linewidth}
    \centering
    \includegraphics[width=\textwidth]{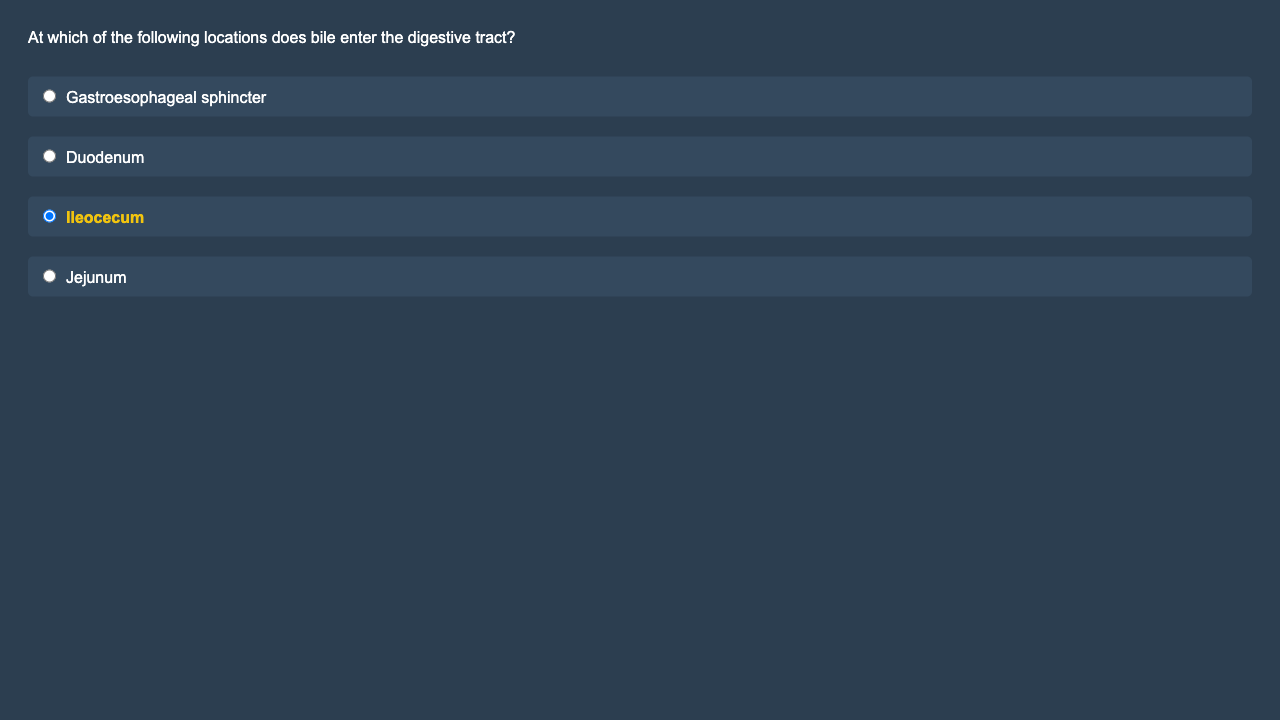}
    \caption{A sample of HTML-rendered vMMLU prompt, option C bias}
    \label{fig:html-vmmlu-b}
\end{minipage}
\end{figure}

\begin{figure}[t!]
\begin{minipage}{0.48\linewidth}
    \centering
    \includegraphics[width=0.8\textwidth]{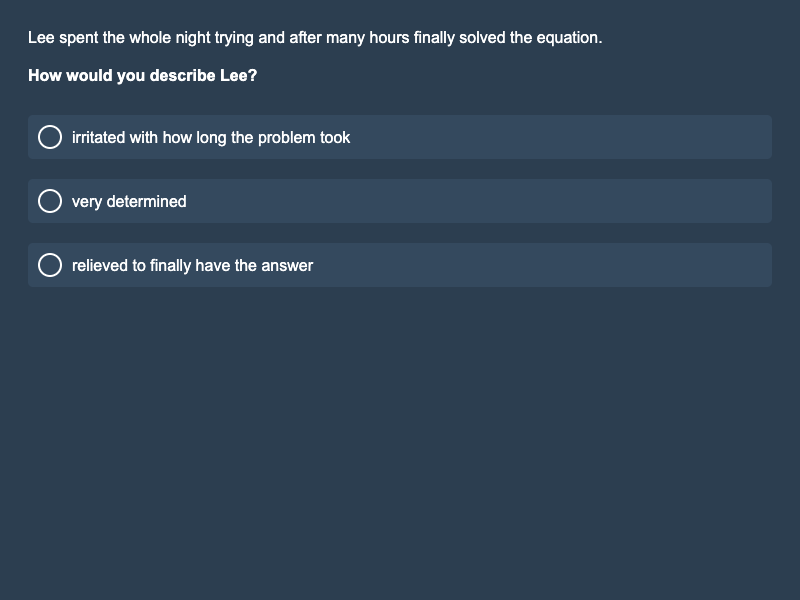}
    \caption{A sample of HTML-rendered vSocialIQa prompt, neutral}
    \label{fig:html-vsocialiqa}
\end{minipage}
\hfill
\begin{minipage}{0.48\linewidth}
    \centering
    \includegraphics[width=0.8\textwidth]{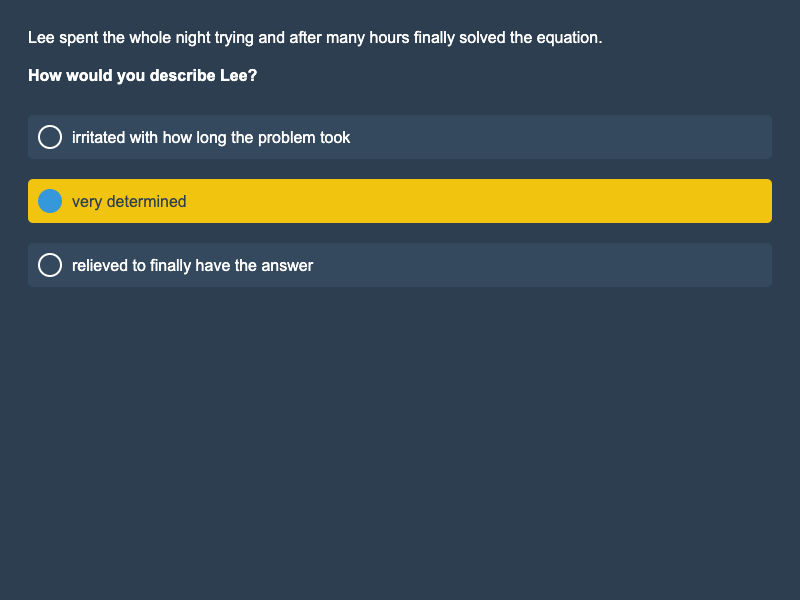}
    \caption{A sample of HTML-rendered vSocialIQa prompt, option B bias}
    \label{fig:html-vsocialiqa-b}
\end{minipage}
\end{figure}

\section{Method}

To empirically investigate the phenomenon of agreeableness bias in multimodal language models, we designed a series of experiments to quantify the influence of pre-marked options on model outputs.

\textbf{Evaluation of Bias.} We measured the shift in the distribution of answers and log probabilities between neutral and biased variations. This involved comparing answer distributions across variations, including the distribution of ground truth answers. We calculated a bias percentage, representing the proportion of answers that changed in both directions. For GPT-4o-mini and LLAVA, we analyzed the shift in the distribution of the top 4 answer token log probabilities across variations.

\textbf{Pre-marked Visual Prompt Generation.} For each neutral prompt in our benchmarks, we generated biased variations by visually pre-marking different answer options. This process created visually distinct versions of the same question, each emphasizing a different answer option.

For the Visual MMLU (vMMLU) benchmark, we produced two types of variation formats:

(1) \textbf{Filled-in bubble with colored text:} We filled in the bubble next to one answer option and colored its text, creating a visual emphasis on that option.

(2) \textbf{Size variation:} We doubled the font size of the biased option without highlighting or bubbling the option.

For the Visual Social IQa (vSocialIQa) benchmark, we also created two types of formats:

(1) \textbf{Bubbled and highlighted text:} We filled in the bubble next to one answer option and highlighted its text in yellow.

(2) \textbf{Web-format style:} We designed a variation resembling a typical web format, featuring a light background, black text for the question and options, and a light blue highlight for one answer option.

\textbf{Visual Prompt Dimensions.} We maintained uniform input dimensions for all prompts and their variations within each benchmark type. For vMMLU, the image prompts were consistently 560 x 640 pixels, and for vSocialIQa, the image prompts were 800 x 600 pixels.

\textbf{vMMLU and vSocialIQa Benchmarks.} We chose these two benchmarks to examine whether the effects of agreeableness bias differ between tasks that require objective information (MMLU) and those that involve nuanced social reasoning (Social IQa) \cite{sap2019socialiqa, hendryckstest2021}. Both variations measure whether there is a significant correlation between shifts in answer distribution and the visually pre-marked options. We chose to experiment with agreeableness bias on both general knowledge and social intelligence tasks to disentangle inconsistencies in factual and social contexts. 
This approach allows us to examine whether the effects of agreeableness bias differ between tasks that require objective information and those that involve nuanced social reasoning. 

Our methodology allows us to quantify the extent to which pre-marked options influence model responses, potentially overriding their prior knowledge or training. By comparing model behaviors across different visual presentations and task types, we aim to provide a comprehensive understanding of agreeableness bias in multimodal AI systems.

\section{Result}
Our experiments on the vMMLU and vSocialIQa benchmarks reveal patterns of agreeableness bias across the tested multimodal language models, with varying degrees of susceptibility among different architectures. Table \ref{tab:model_comparison} summarizes the findings for the vMMLU benchmark.

\begin{table*}[t]
\centering
\resizebox{0.8\textwidth}{!}{
\begin{tabular}{ccccccccc}
\toprule
\multirow{2}{*}{Pre-Marked} & \multirow{2}{*}{Model} & \multicolumn{4}{c}{Response Distribution (\%)} & $\Delta$ Pre & $\Delta$ Not & Score \\
 &  & A & B & C & D & (\%) & (\%) & (\%) \\
\midrule
\multirow{4}{*}{N/A} & claude-haiku & 14.0 & 24.0 & 25.0 & 28.0 & - & - & 64.0 \\
 & gemini-1.5-flash & 17.18 & 18.61 & 27.75 & 36.45 & - & - & 73.02 \\
 & gpt-4o-mini & 18.96 & 34.58 & 28.75 & 17.71 & - & - & 72.08 \\
 & LLaVA & 27.20 & 27.00 & 24.50 & 21.3 & - & - & 50.2 \\
\midrule
\multirow{4}{*}{Option A} & claude-haiku & \textbf{18.0} & 26.0 & 24.0 & 27.0 & +4.0 & 0.0 & 64.0 \\
 & gemini-1.5-flash & \textbf{47.24} & 17.01 & 16.25 & 19.50 & +30.06 & -10.02 & 56.99 \\
 & gpt-4o-mini & \textbf{32.5} & 32.29 & 22.92 & 12.29 & +13.54 & -4.51 & 67.19 \\
 & LLaVA & \textbf{27.00} & 28.80 & 22.60 & 21.60 & -0.20 & -0.47 & 50.2 \\
\midrule
\multirow{4}{*}{Option B} & claude-haiku & 14.0 & \textbf{30.0} & 24.0 & 22.0 & +6.0 & -2.0 & 57.0 \\
 & gemini-1.5-flash & 8.34 & \textbf{55.08} & 16.04 & 20.53 & +36.47 & -12.16 & 62.25 \\
 & gpt-4o-mini & 18.44 & \textbf{45.73} & 24.06 & 11.77 & +11.15 & -3.72 & 63.65 \\
 & LLaVA & 26.80 & \textbf{27.00} & 21.90 & 24.30 & 00.00 & 00.00 & 48.3 \\
\midrule
\multirow{4}{*}{Option C} & claude-haiku & 10.0 & 18.0 & \textbf{45.0} & 21.0 & +20.0 & -5.67 & 60.0 \\
 & gemini-1.5-flash & 7.59 & 8.24 & \textbf{72.73} & 11.44 & +44.98 & -14.99 & 49.84 \\
 & gpt-4o-mini & 16.04 & 26.67 & \textbf{42.5} & 14.79 & +13.75 & -4.58 & 62.92 \\
 & LLaVA & 29.10 & 27.10 & \textbf{22.80} & 21.00 & -1.70 & -0.57  & 52.1 \\
\midrule
\multirow{4}{*}{Option D} & claude-haiku & 7.0 & 18.0 & 17.0 & \textbf{49.0} & +21.0 & -7.0 & 59.0 \\
 & gemini-1.5-flash & 8.59 & 9.42 & 9.32 & \textbf{72.67} & +36.22 & -12.07 & 51.73 \\
 & gpt-4o-mini & 15.02 & 21.48 & 22.21 & \textbf{41.29} & +23.58 & -7.86 & 73.62 \\
 & LLaVA & 26.40 & 28.80 & 22.40 & \textbf{22.40} & -1.10 & -0.37 & 49.3 \\
\bottomrule
\end{tabular}}
\vspace{1mm}
\caption{\textbf{vMMLU Benchmark: Model Performance Comparison on Response Distribution and Changes in Marked and Unmarked Options.} ``$\Delta$ Pre (\%)'' shows the percentage point increase for the pre-marked option compared to the neutral (N/A) condition. ``$\Delta$ Not (\%)'' represents the average change in percentage points for non-pre-marked options, calculated as the sum of changes in the three non-pre-marked options divided by 3. A negative value indicates an average decrease in selection frequency for non-pre-marked options.}
\label{tab:model_comparison}
\end{table*}

\textbf{The response distributions show a consistent trend of shifting towards visually pre-marked options across all tested models.} This shift manifests as an increase in the selection rate for the pre-marked option, accompanied by a corresponding decrease in the selection rates for non-marked options. The magnitude of these shifts varies notably among the models, suggesting differences in visual cues.

\textbf{LLAVA and Claude-haiku demonstrated the most resilience to agreeableness bias}
We observed that LLAVA's performance on vMMLU is at around 50 percent in the neutral setting and does not undergo significant change for pre-marked options. 
Claude-haiku also demonstrated resilience to agreeableness bias, with relatively modest shifts in response distribution. Its largest shift occurred when Option C was pre-marked, resulting in a 20 percentage point increase. Interestingly, Claude-haiku's overall performance remained relatively stable across different pre-marking conditions, with score changes ranging from 0 to -7 percentage points. While Claude-haiku is not immune to agreeableness bias, its impact is limited.

\begin{table*}[t]
\centering
\resizebox{\textwidth}{!}{
\begin{tabular}{ccccccccccccccc}
\toprule
\multirow{3}{*}{Pre} & \multirow{3}{*}{Model} & \multicolumn{6}{c}{Setup A (Webpage Format)} & \multicolumn{6}{c}{Setup B (Yellow Highlight)} \\
\cmidrule(lr){3-8} \cmidrule(lr){9-14}
 &  & \multicolumn{3}{c}{Response (\%)} & $\Delta$ Pre & $\Delta$ Not & Score & \multicolumn{3}{c}{Response (\%)} & $\Delta$ Pre & $\Delta$ Not & Score \\
 &  & A & B & C & (\%) & (\%) & (\%) & A & B & C & (\%) & (\%) & (\%) \\
\midrule
\multirow{3}{*}{N/A} & claude & 23.0 & 30.0 & 47.0 & - & - & 75.0 & 24.0 & 29.0 & 47.0 & - & - & 76.0 \\
 & gemini & 25.39 & 28.67 & 45.94 & - & - & 73.38 & 21.16 & 32.80 & 46.04 & - & - & 76.63 \\
 & gpt-4o & 30.0 & 33.33 & 36.67 & - & - & 86.67 & 25.7 & 35.0 & 39.3 & - & - & 82.7 \\
\midrule
\multirow{3}{*}{A} & claude & \textbf{29.0} & 28.0 & 43.0 & +6.0 & -3.0 & 72.0 & \textbf{66.0} & 17.0 & 17.0 & +42.0 & -21.0 & 52.0 \\
 & gemini & \textbf{55.26} & 23.29 & 21.45 & +29.87 & -14.94 & 67.72 & \textbf{43.57} & 28.51 & 27.91 & +22.41 & -11.21 & 69.48 \\
 & gpt-4o & \textbf{38.0} & 25.33 & 36.67 & +8.0 & -4.0 & 88.0 & \textbf{34.2} & 33.0 & 32.8 & +8.5 & -4.25 & 76.8 \\
\midrule
\multirow{3}{*}{B} & claude & 22.0 & \textbf{37.0} & 41.0 & +7.0 & -3.5 & 74.0 & 3.0 & \textbf{94.0} & 3.0 & +65.0 & -32.5 & 37.0 \\
 & gemini & 12.68 & \textbf{71.57} & 15.75 & +42.9 & -21.45 & 59.0 & 8.91 & \textbf{74.27} & 16.82 & +41.47 & -20.74 & 52.95 \\
 & gpt-4o & 26.67 & \textbf{39.67} & 33.67 & +6.34 & -3.17 & 86.33 & 26.0 & \textbf{44.0} & 30.0 & +9.0 & -4.5 & 77.0 \\
\midrule
\multirow{3}{*}{C} & claude & 16.0 & 22.0 & \textbf{62.0} & +15.0 & -7.5 & 63.0 & 0.0 & 0.0 & \textbf{100.0} & +53.0 & -26.5 & 44.0 \\
 & gemini & 14.72 & 16.36 & \textbf{68.92} & +22.98 & -11.49 & 64.72 & 8.82 & 15.63 & \textbf{75.55} & +29.51 & -14.76 & 65.13 \\
 & gpt-4o & 30.0 & 20.0 & \textbf{50.0} & +13.33 & -6.67 & 76.67 & 19.1 & 24.9 & \textbf{56.0} & +16.7 & -8.35 & 76.9 \\
\bottomrule
\end{tabular}}
\caption{\textbf{vSocialIQa Benchmark: Side-by-Side Comparison of Setup A and B.} Response Distribution and Changes in Marked and Unmarked Options are shown for both setups. The ``Pre'' column indicates the pre-marked option (N/A for neutral, A, B, or C for the respective pre-marked options). ``$\Delta$ Pre (\%)'' shows the percentage point increase for the pre-marked option compared to the neutral (N/A) condition. ``$\Delta$ Not (\%)'' represents the average change in percentage points for non-pre-marked options, calculated as the sum of changes in the two non-pre-marked options divided by 2. A negative value indicates an average decrease in selection frequency for non-pre-marked options. Pre-marked options are in bold. Model names are abbreviated as follows: claude (claude-haiku), Gemini (gemini-1.5-flash), and GPT-4o (gpt-4o-mini).}
\vspace{-3mm}
\label{tab:social_iqa_comparison}
\end{table*}

In contrast, \textbf{Gemini-1.5-flash exhibited the highest susceptibility to agreeableness bias.} It showed substantial shifts in response distribution for all pre-marked options, with increases ranging from 30.06 to 44.98 percentage points. These large shifts were accompanied by more pronounced decreases in non-marked option selection, averaging between -10.02 and -14.99 percentage points. Notably, Gemini-1.5-flash also experienced the most significant performance degradation, with score decreases of up to 23.18 percentage points when Option C was pre-marked. The bias was so pronounced that visual variations could be classified based on the responses alone, with Gemini selecting visually highlighted options A or B about 50 percent of the time and options C or D approximately 70 percent of the time. This pattern suggests a strong influence of visual cues on Gemini-1.5-flash's decision-making at the expense of accuracy. Interestingly, Gemini demonstrated a notable difference in its behavior between blue-centered and vanilla vSocialIQa formats, indicating that the visual presentation of options can significantly impact model responses, potentially mitigating or exacerbating biases.

\textbf{GPT-4o-mini displayed an intermediate level of susceptibility to agreeableness bias.} Its shifts towards pre-marked options, while noticeable, were generally less pronounced than those of Gemini-1.5-flash but more substantial than Claude-haiku's. 
Interestingly, GPT-4o-mini's performance impact varied depending on the pre-marked option, with both slight improvements and decreases observed. 
This variability hints at a complex interaction between visual cues and the model's underlying knowledge or decision-making processes.

\textbf{The strength of the agreeableness bias also varied depending on which option was pre-marked.} Across all models, pre-marking Options C and D generally elicited stronger effects compared to Options A and B. 
This pattern raises questions about potential positional biases or the influence of option ordering on the models' susceptibility to the visual cue.
The observations from the vMMLU benchmark are further nuanced by the results from the vSocialIQa task, as shown in Table \ref{tab:social_iqa_comparison}. 
The vSocialIQa results not only corroborate the presence of agreeableness bias across different task types but also reveal how the effect can be modulated by subtle changes in visual presentation.

\textbf{Visual design modulates agreeableness effects.} The comparison between Setup A and Setup B in the vSocialIQa task demonstrates that the visual presentation of options can dramatically alter the magnitude of agreeableness bias. 
This is most strikingly illustrated by Claude-haiku's performance. 
While it showed resilience in the vMMLU task and in vSocialIQa Setup A, it exhibited extreme susceptibility in Setup B, with shifts of up to 65 percentage points when Option B was pre-marked and a complete 100\% selection of Option C when it was pre-marked. 
This stark contrast underscores the critical role of visual design in multimodal tasks and suggests that model behavior can be highly sensitive to seemingly minor changes in presentation.

\textbf{Consistency of susceptibility varies across models.} Gemini-1.5-flash's high susceptibility to agreeableness bias, observed in the vMMLU task, is consistently evident across both vSocialIQa setups. 
This reinforces the notion that some models may have a more fundamental vulnerability to visual cues, regardless of the specific task or visual presentation. 
In contrast, GPT-4o-mini's intermediate level of susceptibility in vMMLU is mirrored in its relatively stable behavior across both vSocialIQa setups, suggesting a more robust integration of visual and textual information.

\section{Analysis}

Our analysis delves deeper into the patterns of agreeableness bias observed in the results, focusing on changes in token probability and the varying susceptibility of different models across tasks and visual formats.

\begin{figure*}[h]
    \centering
    \textbf{Gpt-4o-mini}
    
    \begin{tabular}{cc}
        \includegraphics[width=0.48\textwidth]{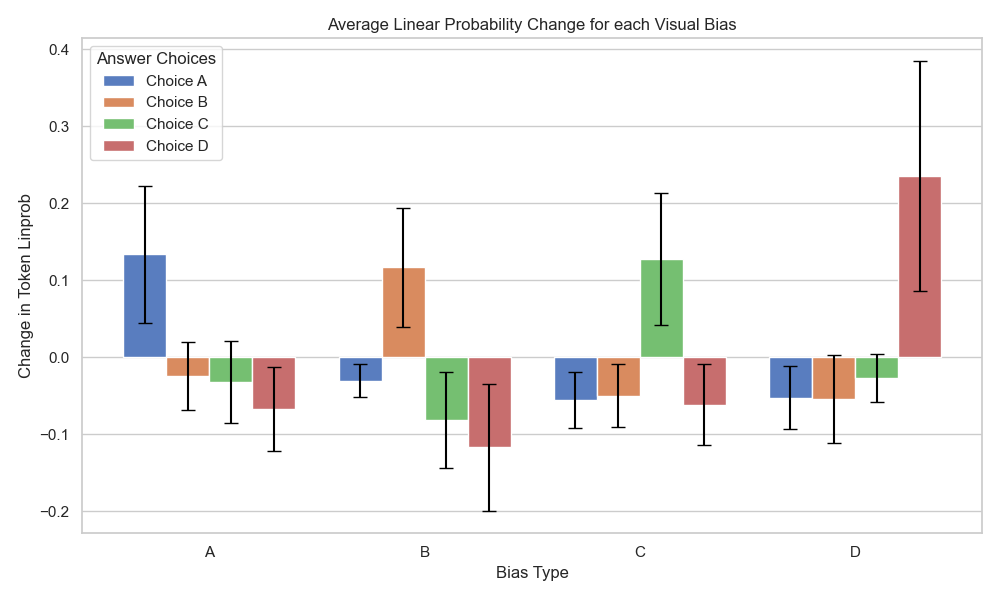} &
        \includegraphics[width=0.48\textwidth]{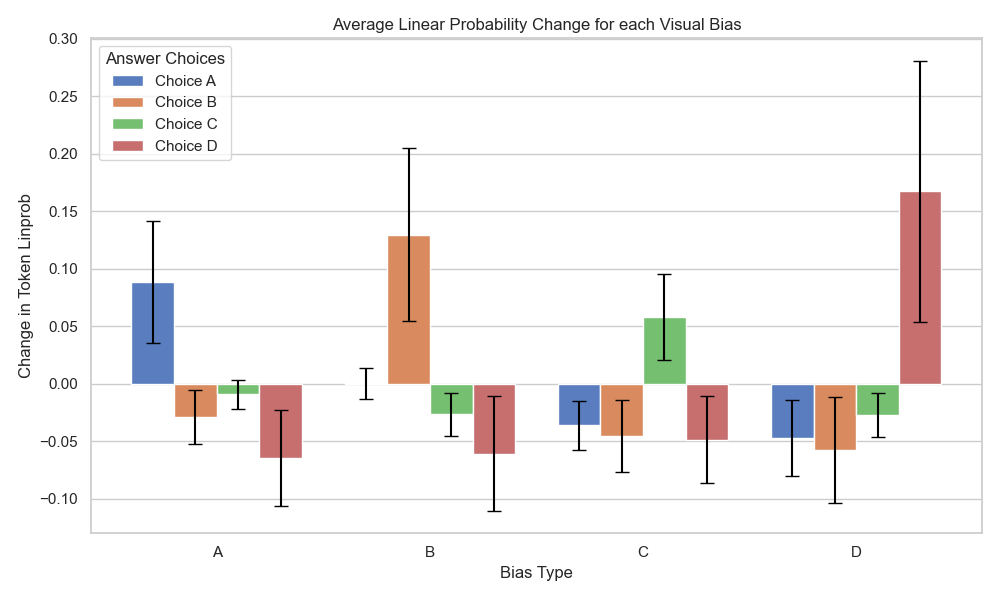} \\
        \includegraphics[width=0.48\textwidth]{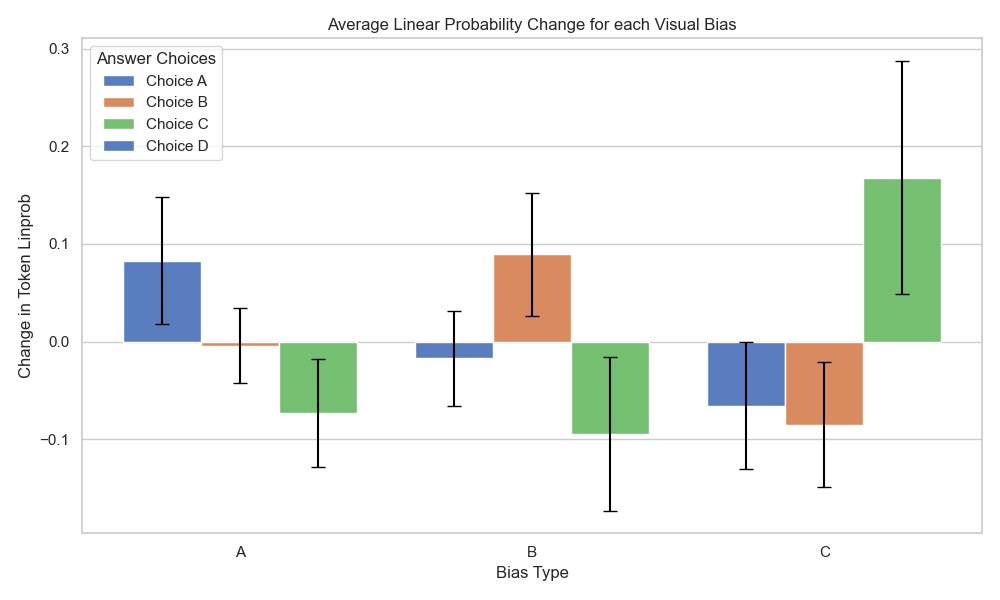} &
        \includegraphics[width=0.48\textwidth]{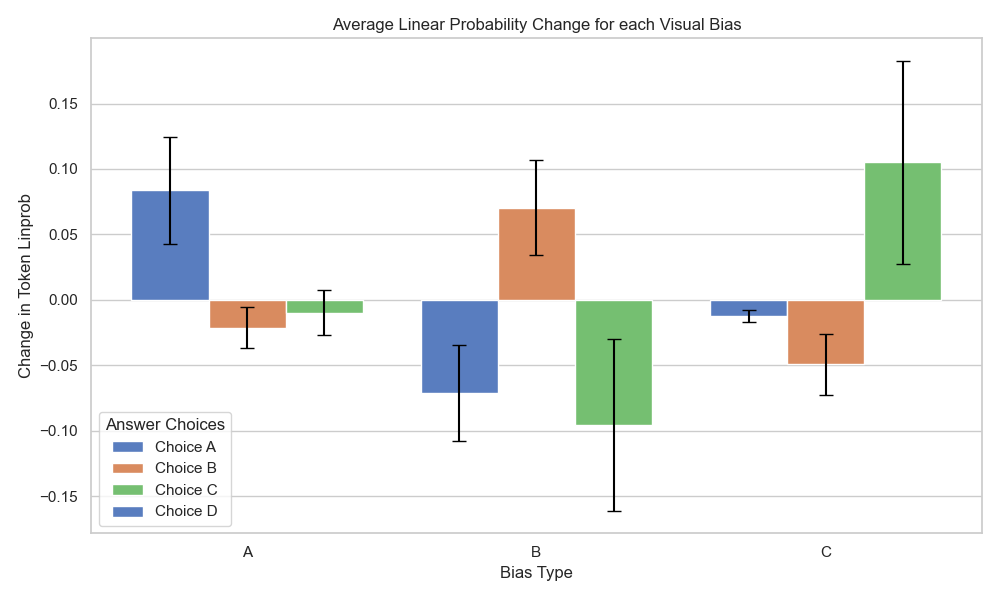}
    \end{tabular}
    \caption{Average change in linear probability between neutral and biased prompts for vMMLU (top row) and vSocialIQa (bottom row). The left column represents highlight bias. The top right plot displays size bias, and the bottom right plot shows highlight bias in a typical webpage format, where black text is highlighted in light blue. The type of bias strongly correlates with increased token probability for the corresponding answer choice.}
    \label{fig:bias-linear}
\end{figure*}

\begin{figure}[t]
\centering
\includegraphics[width=\columnwidth]{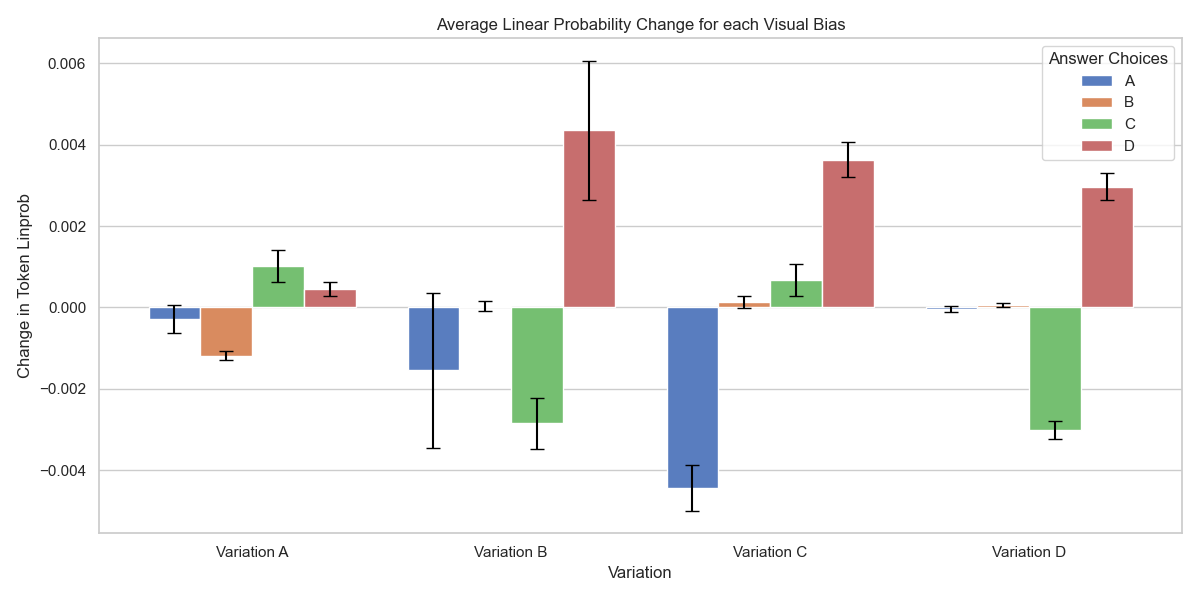}
\caption{Average linear probability change for LLAVA-1.5v-13b across visual bias variations, demonstrating a consistent preference for Option D in vMMLU tasks regardless of the type of visual bias.}
\label{fig:bias-linear-2}
\end{figure}

\textbf{Changes in Token Probability}
We examined the shift in log probability for answer tokens ('A,' 'B,' 'C,' and 'D') across different visual bias conditions. For each bias type, we calculated the change in token probability from neutral to biased conditions, averaged across prompts. To avoid unintended scaling effects, we first converted the probabilities back to linear probability before calculating deltas. As shown in Figure \ref{fig:bias-linear}, for GPT-4o-mini, the bias type strongly correlates with increased token probability for the corresponding answer choice. This indicates that the model is most likely to increase the probability of a given token when the visual information suggests that the token is the answer. Interestingly, LLAVA-1.5v-13b exhibited a different pattern of bias, as shown in Figure \ref{fig:bias-linear-2}. Instead of showing a stronger preference for the biased option, LLAVA consistently showed a stronger preference for option D across all biased variations, with the exception of variation A, where options C and D were tied for positive delta.

\textbf{Claude-haiku exhibits varying degrees of susceptibility to visual bias across different benchmarks and formats.} 
In the vMMLU benchmark, the model shows minimal bias, with only slight increases in responses for visually emphasized options. The effect becomes more pronounced in the webpage-formatted Social IQa, where emphasized options receive moderately increased selection. However, the most dramatic impact is observed in the vanilla vSocialIQa format, where the bias is extreme—reaching 100 percent selection for visually emphasized option C. This progression suggests that Claude-haiku's sensitivity to visual cues intensifies as the visual emphasis becomes more pronounced, with the vanilla vSocialIQa format eliciting the strongest bias response. The varying performance across these tests underscores the importance of considering visual formatting in evaluating language model behaviors.

\section{Related Work}

Our study on agreeableness bias in multimodal language models builds upon several key research areas. Recent advancements in multimodal AI systems have expanded the capabilities of models to process and integrate information from various modalities \citep{wu2024semanticequivalencetokenizationmultimodal, ge2024worldgptempoweringllmmultimodal, li2024cumoscalingmultimodalllm, li2024tokenpackerefficientvisualprojector}. This progress has been accompanied by growing concerns about biases in AI, including both social and modality-specific biases \citep{luo2024bigbench, alabdulmohsin2024clipbiasusefulbalancing, lu2024revisitingmultimodalllmevaluation, adewumi2024fairnessbiasmultimodalai, chen2024quantifyingmitigatingunimodalbiases}. 
Visual attention mechanisms in AI systems have been studied extensively, often drawing parallels with human visual processing \citep{cao2024visualcognitiongaphumans}. Evaluation methodologies for multimodal AI systems have evolved to address the complexities of assessing performance across different modalities \citep{ye2024mmspubenchbetterunderstandingspurious}. Simultaneously, ethical considerations in AI development have gained prominence, focusing on the potential impacts of AI biases in real-world applications \citep{amirloo2024understandingalignmentmultimodalllms}.

\section{Limitations and Future Work}

There are several limitations to this work:

Although we investigate and analyze a few of the most well-known proprietary and open-source models, the generalizability of our findings could be enhanced by including a broader range of state-of-the-art models. Future work should aim to test these concepts across an even more diverse set of architectures and configurations. Since multimodal prompt engineering and agreeableness bias are relatively new concepts, our primary focus has been on measuring the bias rather than proposing specific applications or effective jailbreak mitigation strategies. Further research is needed to develop practical interventions and assess their effectiveness in real-world scenarios.

The use of token probability delta as a novel metric for calculating bias in machine learning models is still in its early stages. It is not yet entirely clear whether systematic bias in multimodal machine learning models is inherently additive or subtractive, and this remains an area for further empirical and theoretical investigation. Modalities other than vision and text have not been explored in this study. Future research should consider extending the analysis to include other modalities, such as audio or sensor data, to determine whether agreeableness bias or similar biases are present across different types of multimodal inputs.

Our experiments primarily focus on visual biases within the context of multiple-choice benchmarks. The extent to which these findings translate to more complex, free-text generation tasks or other forms of human-AI interaction remains unexplored and could be an avenue for future research. While our study addresses the phenomenon of agreeableness bias, we have not fully explored the potential interactions between visual biases and other types of biases (e.g., social or cognitive biases) present in multimodal models. Understanding these interactions could provide a more comprehensive picture of bias in AI systems.

\section{Conclusion}

Our study on agreeableness bias in multimodal language models reveals a complex landscape of model behaviors and biases. We found that the susceptibility to visual cues varies significantly across different model architectures, task types, and visual presentation formats. 
Our findings challenge simplistic assumptions about multimodal information integration in AI systems and raise important questions about the reliability and consistency of model outputs. The observed agreeableness bias effects underscore the need for careful consideration of visual elements in the design and deployment of multimodal AI systems, particularly in critical decision-making contexts.

\bibliography{custom}

\end{document}